%% file: Main.tex
\documentclass{article}
\usepackage{ijcai23}

\usepackage{times}
\usepackage{soul}
\usepackage{url}
\usepackage[hidelinks]{hyperref}
\usepackage[utf8]{inputenc}
\usepackage[small]{caption}
\usepackage{graphicx}
\usepackage{amsmath}
\usepackage{amsthm}
\usepackage{booktabs}
\usepackage{algorithm}
\usepackage{algorithmic}
\usepackage[switch]{lineno}
\usepackage{multirow}

\usepackage[table]{xcolor}

\title{When Layers Play the Lottery, all Tickets Win at Initialization}

\author{
Artur Jordao$^1$, George Corrêa de Araújo$^2$, Helena de Almeida Maia$^2$ \And Helio Pedrini$^2$
\affiliations
$^1$Escola Politécnica, Universidade de São Paulo\\$^2$Institute of Computing, University of Campinas\\
}

\begin{document}

\maketitle

\input{Sections/Abstract}

\input{Sections/Introduction}
%
\input{Sections/RelatedWork}
\input{Sections/Methodology}
\input{Sections/Experiments}
\input{Sections/Conclusions}
%
\input{Sections/Acknowledgments.tex}

\bibliographystyle{named}
\bibliography{refs}
\input{Sections/Appendix}

\end{document}

%% file: Sections/Abstract.tex
\begin{abstract}
Pruning is a standard technique for reducing the computational cost of deep networks. 
Many advances in pruning leverage concepts from the Lottery Ticket Hypothesis (LTH).
%
%
LTH reveals that inside a trained dense network exists sparse subnetworks (tickets) able to achieve similar accuracy (i.e., win the lottery -- \emph{winning tickets}). 
%
%
%
Pruning at initialization focuses on finding winning tickets without training a dense network.
%
Studies on these concepts share the trend that subnetworks come from weight or filter pruning. 
In this work, we investigate LTH and pruning at initialization from the lens of layer pruning.
First, we confirm the existence of winning tickets when the pruning process removes layers.
Leveraged by this observation, we propose to discover these winning tickets at initialization, eliminating the requirement of heavy computational resources for training the initial (over-parameterized) dense network.
%
Extensive experiments show that our winning tickets notably speed up the training phase and reduce up to 51\% of carbon emission, an important step towards democratization and green Artificial Intelligence.
Beyond computational benefits, our winning tickets exhibit robustness against adversarial and out-of-distribution examples.
%
Finally, we show that our subnetworks easily win the lottery at initialization while tickets from filter removal (the standard structured LTH) hardly become winning tickets.
\end{abstract}

%% file: Sections/Introduction.tex
\section{Introduction}\label{sec:introduction}
The Lottery Ticket Hypothesis (LTH) conjectures that (pre-trained) dense networks contain sparse subnetworks capable of obtaining the same accuracy when trained from the original initialization of their dense counterpart~\cite{Frankle:2019}. Subnetworks (tickets) that satisfy this property are referred to as \emph{winning tickets}.
Many advances emerge from the LTH, for example, we can reduce the computational cost of learning a dense network by replacing it with a sparse subnetwork during ~\cite{You:2020,Chen:2022} or before the course of training~\cite{Lee:2019,Wang:2020,Tanaka:2020}.
In the context of adversarial attacks, we can increase the robustness of dense networks by considering their sparse versions~\cite{Diffenderfer:2021,Liu:2022,T:2022}.
\begin{figure}[!t]
	\centering
	\includegraphics[width=\columnwidth]{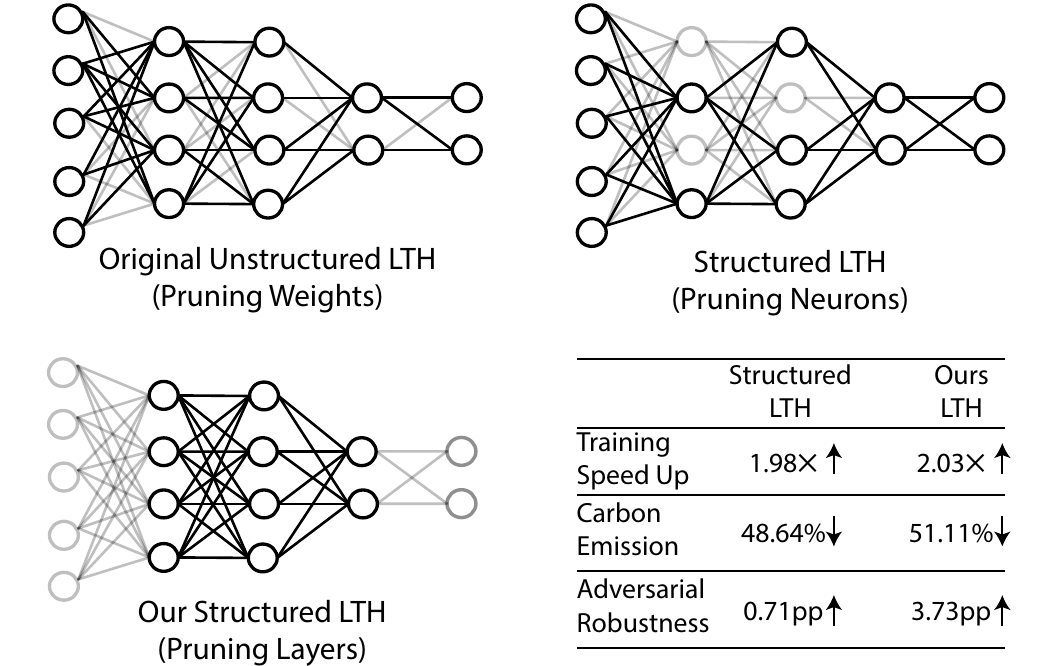}
	\caption{Lottery Ticket Hypothesis (LTH) views according to the structure (weights, neurons/filters or layers) the pruning is eliminating (transparent regions).
	Top-left. Original unstructured LTH: the pruning removes weights and yields unstructured tickets; thereby, the tickets only provide practical benefits on specialized frameworks for sparse computations. Top-right. Structured LTH: the pruning eliminates neurons/filters. In this setting, the tickets are structured and promote computational advantages to standard deep learning frameworks.
	Bottom-left: Ours structured LTH: the pruning eliminates entire layers, encouraging additional performance gains since it decreases the sequential processing (latency). Bottom-right. The highest gain (the higher, the better) obtained by a winning ticket regarding its dense counterpart.
	Our winning tickets successfully emerge at initialization, which means we can discover efficient subnetworks without training a dense network. 
	In this direction, we can considerably speed up the learning phase by replacing a dense network with its sparse version before training begins. Our winning tickets also exhibit robustness against adversarial attacks.
	%
}
	\label{fig:teaser}
\end{figure}

The traditional LTH seeks winning tickets by removing (i.e., pruning) weights of a dense network and assigning the survival weights to their original initialization~\cite{Frankle:2019} -- weight rewinding.
%
%
Variants of this mechanism propose relaxing the weight-rewinding constraint and enabling subnetworks to inherit weights from different training epochs~\cite{Frankle:2020,Renda:2020} and even from random (re) initialization~\cite{Liu:2019}.
%
%
Another variant in the traditional LTH is the replacement of weight pruning (unstructured) by filter pruning (structured)~\cite{Liu:2019,Renda:2020,You:2020}. For example, Prasanna et al.~\shortcite{Prasanna:2020} studied the LHN on structured pruning by removing self-attention modules (i.e., heads from multi-head attention layers~\cite{Vaswani:2017}).
One practical drawback of unstructured pruning is that existing deep learning frameworks (e.g., TensorFlow and Pytorch) do not support sparse tensor computations. Hence, in order to obtain performance gains, this family of pruning requires specialized frameworks or hardware (Nvidia A100 GPU) for optimizing sparse computations~\cite{Han:2016,Niu:2020,Zhou:2021}.
%
%

Regardless of the structure, we can categorize LTH according to the phase in which the pruning algorithm picks the subnetworks (tickets): after, during, or before training. 
The first class trains a dense network and, then, extracts a subnetwork according to an importance criterion (pruning criterion), such steps compose the original LTH~\cite{Frankle:2019}.
The second prunes a dense network during the course of training, obtaining a subnetwork when the learning phase is over~\cite{You:2020,Chen:2022}. Finally, pruning before training constitutes a relatively recent form of pruning named \emph{pruning at initialization}~\cite{Lee:2019,Tanaka:2020,Wang:2020}. 
The idea behind this category is to find subnetworks prior to training, which means picking subnetworks using the randomly initialized parameters (without any update) to guide the pruning algorithm.
Many efforts have been put into pruning at initialization as it provides all benefits of sparse subnetworks without spending computational resources to train a dense network (a process required by the other categories).

%
Importantly, all pruning categories in LTH share the same characteristic: the pruning eliminates small structures such as weights or filters. Outside the context of LTH, previous works have demonstrated notable benefits of removing entire layers instead of small structures. 
%
In particular, this type of structured pruning promotes additional performance gains since it reduces the sequential processing (latency) of a network~\cite{Li:2021:CVPR,Zhou:2022}. 
%
Hence, many studies have focused on removing layers instead of other structures~\cite{Veit:2020,Fan:2020,Zhou:2022,Zhang:2022}.
%
Unfortunately, none of these efforts have been done in the direction of LTH and pruning at initialization. To bridge this gap, we take a step towards understanding the behavior of LTH when the pruning process removes layers. In this setting, we first verify the existence of winning tickets. Then, we propose a systematic strategy for discovering winning tickets before training, which means finding sparse subnetworks (from layer removal) at initialization that match the predictive ability of their dense equivalent.
%
%

\noindent
\textbf{Contributions.} We list the following key contributions. 
First, we demonstrate the existence of winning tickets -- sparse subnetworks that obtain the same accuracy as their dense equivalent -- when the pruning process takes into account the removal of layers. 
From a practical perspective, such a contribution enables winning tickets to be more efficient in terms of memory consumption, inference time (latency) and carbon emission, as removing layers provides superior computational benefits than standard forms of pruning employed in LTH (see Figure~\ref{fig:teaser}).
Second, we successfully find winning tickets from layer removal at initialization without any training. This contribution plays a role in low-cost and energy-efficient training, as we can replace the learning of a dense network with its sparse version. In contrast to previous LTH studies~\cite{Frankle:2020,Renda:2020}, this contribution eliminates the requirement of heavy computational resources for learning the initial (over-parameterized) dense network.
%
Finally, we show that this novel family of subnetworks shares some properties of the standard structured LTH (filter pruning), for example, they emerge early at training. 
On the other hand, we observe that subnetworks from filter pruning exhibit poor performance when applied to shallow (underparametrized) networks and before training, which is aligned with previous evidence~\cite{Liu:2022}. In other words, tickets from filter pruning hardly become winning tickets. 
However, our tickets (layer removal) easily win the lottery, raising the question of whether the eliminated structure plays a role in the LTH and pruning at initialization.
%
%
%

According to our analysis, winning tickets from layer pruning are accurate and robust to many settings such as the pruning criteria and density, and weight rewinding. When found at initialization, these winning tickets speed up the training time by up to $2\times$ and save notable FLOPs and memory consumption. 
On a specific pruning criterion, all tickets outperform the predictive ability of their dense counterpart -- all tickets become winning tickets. Additionally, our winning tickets reduce up to $51\%$ of carbon emission during the training phase, an important step towards democratization and sustainability of Artificial Intelligence (AI) -- green AI.
%
%
%
As suggested by previous works~\cite{Liu:2022,T:2022,Chen:2022}, we also evaluate our winning tickets on adversarial images and out-of-distribution examples using CIFAR-C~\cite{Hendrycks:2019} and CIFAR-10.2~\cite{Lu:2020} datasets, respectively. 
On the standard clean training (i.e., without any defense mechanism), tickets from layer pruning achieve higher robustness than their dense equivalent; thus, confirming their suitability for safety-critical applications. For reproducibility purposes, our code is publicly available: \emph{https://github.com/arturjordao/LayerLottery}.
%

%% file: Sections/RelatedWork.tex
\section{Related Work}\label{sec:related}
Frankle et al.~\shortcite{Frankle:2019} introduced the Lottery Ticket Hypothesis and observed that subnetworks from a dense network can obtain similar accuracy since trained from the same original initialization. 
%
Later, Frankle et al.~\shortcite{Frankle:2020} and Renda et al.~\shortcite{Renda:2020} confirmed that rewinding the weights to other epochs of the training, instead of the original initialization, enables subnetworks to match the accuracy of their dense counterpart. 

The work by Achille et al.~\shortcite{Achille:2019} investigated the sensibility of networks during training. The authors observed that perturbations, such as blur and random label, affect less the training dynamic when applied in early epochs. By considering pruning as a form of perturbation (e.g., due to the removal of structures), such evidence reinforces the superior performance of subnetworks rewound to early training epochs, as argued by subsequent studies~\cite{You:2020}.
%
While weight rewinding lies at the heart of LTH, Liu et al.~\shortcite{Liu:2019} demonstrated that the winning tickets exist even on a random initialization regime.
%
Throughout this study, we show that our winning tickets (resulting from layer pruning) emerge from different rewinding epochs, random initialization and before training, with the latter achieving the best gains in terms of accuracy and computational performance.


Liu et al.~\shortcite{Liu:2021} suggested that the existence of winning tickets correlates with the transition from the initial and final parameters of the dense network.
From a different perspective, Paul et al.~\shortcite{Paul:2022} showed that data quantity and quality play a role in LTH. More concretely, the authors observed that training on easy or on a small fraction of randomly chosen data promotes a fine initialization to dense networks such that their subnetworks become winning tickets.
%
%
%
In line with Liu et al.~\shortcite{Liu:2021}, You et al.~\shortcite{You:2020} proposed drawing sparse subnetworks early at training by identifying when the magnitude of the weights becomes stable (i.e., suffer small changes) during the course of training. After locating a subnetwork, named early bird, the authors replaced the training of the dense with the subnetwork; thus leading to faster and lower-cost training. Built upon this idea, Chen et al.~\shortcite{Chen:2022} proposed to find early birds to reduce the cost of adversarial training. However, the authors identify subnetworks by pruning weights (unstructured) while You et al.~\shortcite{You:2020} focus on removing filters (structured pruning). Interestingly, the results by Chen et al.~\shortcite{Chen:2022} corroborate the observation of previous works~\cite{Diffenderfer:2021,Liu:2022}, which state that pruning not only reduces the computational cost but also promotes robustness against adversarial images. Additionally, T et al.~\shortcite{T:2022} confirmed that winning tickets generalize better than dense networks in limited-data regimes and out-of-distribution scenarios.

Beyond computational benefits, You et al~~\shortcite{You:2020} and Chen et al.~\shortcite{Chen:2022} showed that winning tickets drawn early at training obtain higher accuracy. 
From the LTH rewinding perspective, this means that resetting the weights of subnetworks to early epochs of the dense network allow subnetworks to achieve higher predictive ability, hence, they are more probable to become a winning ticket. Our results are aligned with these findings: subnetworks from layer pruning achieve superior accuracy when rewound to early epochs. In contrast to these works, we also propose to identify winning tickets at initialization (prior to training). Informally speaking, our subnetworks are the earliest as possible.
%
%

Closely related to LTH, several efforts have been put into pruning at initialization. In this strategy, the pruning algorithm removes unimportant structures before any training~\cite{Lee:2019,Wang:2020,Tanaka:2020}. The study by Frankle et al.~\shortcite{Frankle:2021} pointed out the difficulties and inherent properties of many pruning at initialization strategies. 
The authors showed that this family of strategies is less effective than standard pruning (i.e., LTH) in terms of both accuracy and sparsity. Jorge et al.~\shortcite{Jorge:2021} corroborate such findings; further, they observed that pruning at initialization is no better than random pruning on high sparsity regimes. 
%
Liu et al.~\shortcite{Liu:2022} suggested an opposite behavior: pruning at initialization works well on deep and wide networks, even randomly selecting the unimportant structures. Additionally, the authors showed that employing a layer-wise pruning density yields more accurate winning tickets.
We observe that the evidence by Liu et al.~\shortcite{Liu:2022} holds when pruning takes into account filters; however, we successfully find winning tickets at initialization on both shallow (i.e., ResNet32) and deep architectures (i.e., ResNet56), which suggest an architecture-agnostic form of discovering winning tickets prior to training. Though it looks like a counter-intuitive phenomenon, there is a body of studies confirming the benefits of removing layers instead of filters~\cite{Zhang:2022,Zhou:2022}.
In summary, our work differs from previous studies on LTH and pruning at the initialization in terms of the structure we focus on removing -- layers. 

To the best of our knowledge, the idea behind pruning layers dates back to 2016, when Veit et al.~\shortcite{Veit:2016} and Huang et al.~\shortcite{Huang:2016} demonstrated that residual-based architectures exhibit no degradation when we remove some of their layers. Later, Dong et al.~\shortcite{Dong:2021} confirmed that self-attention architectures also share similar behavior.
Since these pioneer studies, many works have proposed either removing layers statically or dynamically. It is important to mention the difference between these categories of layer pruning. In the former, the pruning algorithm eliminates layers in the same way as weight and filter pruning does, yielding a permanently shallower network~\cite{Zhang:2022,Zhou:2022}. Differently, the rationale behind the dynamic layer pruning consists of skipping (i.e., deactivating) some layers according to the input sample the network receives~\cite{Han:2022}.
%
Our work belongs to the first category of layer pruning. In particular, due to its dynamic nature, we believe it is impractical to study LTH and pruning at the initialization in the second group.

%% file: Sections/Methodology.tex
\section{Preliminaries and Problem Statement}
\noindent
\textbf{Definitions.} Let $\mathcal{F}$ and $\mathcal{F'}$ be a dense and sparse (subnetwork) network, respectively. The latter is a version of $\mathcal{F}$ without some structures (neurons, filters or layers), meaning $\mathcal{F}$ after some pruning process. Assume $\theta_i$ the parameters from $\mathcal{F}$, in which the subscript $i$ indicates the weights at the training epoch $i$. We indicate the randomly initialized weights and the ones after the training stage as $\theta_0$ and $\theta_n$, in this order.

\noindent
\textbf{Pruning Algorithm.}
%
A pruning algorithm identifies and removes unimportant structures composing a network. For this purpose, it measures the importance of each structure according to an importance criterion $c$. Let $S$ be a set of (sorted) scores that indicates the importance of each structure of $\mathcal{F}$. Given $S$, the pruning algorithm removes the least important structures in order to satisfy a pruning density $p$ (i.e., the percentage or number of structures removed.). In the literature on pruning, it is common to mask the eliminated structures with zeros values. In contrast, we indeed remove the structures (layers) to achieve practical performance gains without needing specialized frameworks or hardware for sparse computations. We refer interested readers to Appendix~\ref{sec:app_technicaldetails} for additional details about this technical process.

\noindent
\textbf{Lottery Ticket Hypothesis.} Following the previous definitions, the original Lottery Ticket Hypothesis (LTH)~\cite{Frankle:2019} states that inside $\mathcal{F}$ exists sparse subnetworks $\mathcal{F'}$ able to achieve similar (or, ideally, superior) accuracy since trained from the identical initialization $\theta_0$. In this definition, the subnetworks are named tickets and the ones that satisfy such property are named winning tickets. In other words, a winning ticket is a subnetwork with the same predictive ability (i.e., accuracy) as its dense equivalent. In practice, we can define the existence of winning tickets in terms of 
\begin{equation}\label{eq:LTH}
	Accuracy(\mathcal{F'})+\xi \geq Accuracy(\mathcal{F}), \xi \geq 0.
\end{equation}
As we mentioned before, variants of LTH enable $\mathcal{F'}$ subnetworks to inherit
weights from different training epochs (i.e., $\theta_i$ with $i>0$)~\cite{Frankle:2020,Frankle:2021}. This step composes the weight rewinding process. Algorithm~\ref{alg::pruning} summarizes the steps of our LTH (layer pruning).

From the above description, we highlight two observations. (i) In Equation~\ref{eq:LTH}, $\xi$ enables controlling (relaxing) how much the accuracy of a ticket can differ from its dense network and still be considered a winning ticket, where common values are one percentage point or one standard deviation~\cite{Chen:2020:NeurIPS,Frankle:2021}.
%
%
%
(ii) Before extracting a potential winning ticket, we need to train $\mathcal{F}$ to completion (i.e., training for $n$ epochs to obtain $\theta_n$) -- see the first input in Algorithm~\ref{alg::pruning}.

\noindent
\textbf{Pruning at Initialization.} This relatively recent category of pruning estimates which structures to remove before training. Technically speaking, it focuses on discovering subnetworks using the random initialization ($\theta_0$) of $\mathcal{F}$ to guide the pruning algorithm. From the lens of LTH, pruning at initialization aims at yielding winning tickets without training $\mathcal{F}$. 
\begin{algorithm}[!t]
	\caption{LTH Removing Layers of Deep Networks}
	\label{alg::pruning}
	\begin{algorithmic}
		\STATE {\bfseries Input:} Convolutional Network $\mathcal{F}$ trained on $n$ epochs
		\STATE {\bfseries Input:} Weight Rewind $\theta_i$\\
		\STATE {\bfseries Input:} Pruning Criterion $c$\\
		\STATE {\bfseries Input:} Pruning Density $p$\\
		\STATE {\bfseries Output:} Subnetwork (Ticket) $\mathcal{F'}$
		\STATE $S \leftarrow c(\mathcal{F}, \theta_n)$ $\triangleright$ Assigns importance for each layer
		\STATE $I \leftarrow p$ unimportant layers based on $S$
		\STATE $\mathcal{F'} \leftarrow \mathcal{F} \setminus I $ $\triangleright$ Removes the layers indexed by $I$ 
		\STATE Set the weights of $\mathcal{F'}$ as $\theta_i$ $\triangleright$ Weight rewinding
		\STATE Train $\mathcal{F'}$ via standard training for $n-i$ epochs
	\end{algorithmic}
\end{algorithm}

Algorithm~\ref{alg::early_ticket} summarizes the steps of our pruning at initialization. In Algorithm~\ref{alg::early_ticket}, it is important to observe that it receives an untrained network, whereas in our standard LTH (Algorithm~\ref{alg::pruning}) the input is a trained network.
%

\noindent
\textbf{Research Questions.} From the above definitions, our research questions are the following. (i) We ask if there are winning tickets when $\mathcal{F'}$ derives from a \emph{layer-pruning} process.
%
In other words, we investigate if Equation~\ref{eq:LTH} holds when the pruning removes layers.
We confirm that the answer is positive; thus, raising our second question: (ii) Is it possible to discover such winning tickets (from layer pruning) at initialization? Formally, is it possible to extract an $\mathcal{F'}$ that satisfies Equation~\ref{eq:LTH} without training $\mathcal{F}$?
Answering this enables us to avoid the computationally expensive training of a dense network by replacing it directly with its sparse version before training begins.
Overall, both questions focus on analyzing Equation~\ref{eq:LTH} after performing Algorithm~\ref{alg::pruning} and Algorithm~\ref{alg::early_ticket}.
\begin{algorithm}[!t]
	\caption{Winning Tickets at Initialization}
	\label{alg::early_ticket}
	\begin{algorithmic}
		\STATE {\bfseries Input:} Untrained Convolutional Network $\mathcal{F}$
		\STATE {\bfseries Input:} Pruning Criterion $c$ $\triangleright$ SNIP or GraSP\\
		\STATE {\bfseries Input:} Pruning Density $p$\\
		\STATE {\bfseries Output:} Subnetwork (Ticket) $\mathcal{F'}$
		\STATE $S \leftarrow c(\mathcal{F}, \theta_0)$ $\triangleright$ Assigns importance for each layer
		\STATE $I \leftarrow p$ unimportant layers based on $S$
		\STATE $\mathcal{F'} \leftarrow \mathcal{F} \setminus I $ $\triangleright$ Removes the layers indexed by $I$ 
		\STATE Train $\mathcal{F'}$ via standard training for $n$ epochs
	\end{algorithmic}
\end{algorithm}

%% file: Sections/Experiments.tex
\section{Experiments}\label{sec:experiments}
\textbf{Experimental Setup.} 
We conduct our experiments on CIFAR-10~\cite{cifar} using the ResNet32 and ResNet56 networks~\cite{He:2016}. Such settings are common choices for LTH and general pruning evaluation~\cite{Blalock:2020}.
In our LTH, we employ the learning rate rewinding scheme suggested by Renda et al.~\shortcite{Renda:2020}, as it leads to better results than the one proposed in the original LTH~\cite{Frankle:2019}.
%
In contrast to most LTH studies, we evaluate the LTH on multiple criteria instead of employing only the $\ell_1$-norm criterion (the iterative magnitude pruning --- IMP -- originally proposed by Frankle et al.~\shortcite{Frankle:2019}). The reason for this choice is to assess the existence of winning tickets on multiple pruning settings. Additionally, layers at different depths exhibit different magnitudes, hence, it is unfeasible to compare the $\ell_1$-norm from multiple layers (see Appendix~\ref{sec:l1_histogram}). Specifically, we employ the pruning criteria proposed by Lin et al.~\shortcite{Lin:2020} (HRank), Tan and Motani~\shortcite{Tan:2020} (expectABS), Jordao et al.~\shortcite{Jordao:2020} (PLS) and Luo and Wu~\shortcite{Luo:2020} (KL). In addition, we use the commonly used random criterion.

We report (in percentage points -- pp) the improvement and decline of the subnetworks (tickets) with respect to their dense counterpart using the symbols (+) and (-), in this order. Thereby, the symbol (+) stands for winning tickets since these networks match or outperform the accuracy of their dense equivalent. 
%
%
Since our pruning process considers removing layers, it cannot be evaluated on VGG-like architectures (plain and non-residual networks) due to technical and theoretical details. We refer interested readers to Appendix~\ref{sec:app_technicaldetails} for additional information.
%

%
Unless stated otherwise, the term subnetworks indicate networks yielded by the process of removing layers, which is the scope of our work.
In this setting, the pruning density $p$ stands for the number of layers removed, e.g., $p=1$ indicates that the pruning removed 1 residual block. (In ResNet32-56 as well as its shallower and deeper variations, one residual block (layer) corresponds to two convolutional layers.)
\begin{table}[!tb]
	\centering
	\renewcommand{\arraystretch}{1.2}
	\caption{Predictive ability of the tickets when rewinding their weights to $i$-th training epoch ($\theta_i$) of the dense network (ResNet32). For each criterion, we highlight in bold and underline the rewind epoch that leads to the top-1 and top-2 best results, respectively.}
	\label{tab:rewind}
\begin{tabular}{ccccc}
	& $\theta_{0}$                 & $\theta_{25}$        & $\theta_{50}$                  & $\theta_{75}$       \\ \hline
	\multicolumn{1}{c|}{HRank}     & (-) 0.01          & \textbf{(+) 0.21}    & \underline{(+) 0.10} & (+) 0.12 \\
	\multicolumn{1}{c|}{expectABS} & \textbf{(+) 0.33} & \underline{(+) 0.13} & (-) 0.48             & (-) 0.62 \\
	\multicolumn{1}{c|}{PLS}       & (+) 0.06          & \textbf{(+) 0.31}    & \underline{(+) 0.10} & (+) 0.10 \\
	\multicolumn{1}{c|}{Random}    & (+) 0.38          & (+) 0.15             & \underline{(+) 0.26} & (-) 0.07 \\
	\multicolumn{1}{c|}{KL}        & \textbf{(+) 0.74} & \underline{(+) 0.46} & (-) 0.03             & (-) 0.17 \\ \hline
\end{tabular}
\end{table}

\noindent
\textbf{Existence of Winning Tickets and Weight Rewinding.} 
Our first experiment verifies if subnetworks become winning tickets when pruning removes layers. One of the most important facts to a subnetwork becoming a winning ticket is the epoch we rewind its weights (weight rewinding), which means the $\theta_i$ that the subnetwork inherits from its dense version.

Previous studies on LTH confirmed that inheriting weights from early epochs enables subnetworks successfully become winning tickets~\cite{You:2020,Frankle:2021}. From a practical perspective, resetting the weights to these epochs requires more training iterations to completion (see the last step in Algorithm~\ref{alg::pruning}). In this experiment, we evaluate the effect of setting the weights of subnetworks from layer pruning to different epochs. In other words, we study the behavior of subnetworks when they inherit different $\theta_i$.

Table~\ref{tab:rewind} summarizes the results. From this table, we highlight the following observations. First, rewinding subnetworks to the same (random) initialization of the dense network, $\theta_0$, enables all subnetworks to become winning tickets (except HRank which exhibited a negligible drop). This means that, rewinding to $\theta_0$, all subnetworks satisfy Equation~\ref{eq:LTH}.
Second, most subnetworks achieve the lowest accuracy rewinding after $50$ epochs. 
This aligns with the findings of Achille et al.~\shortcite{Achille:2019} and You et al.~\shortcite{You:2020}: rewinding to late epochs, the subnetworks obtain lower accuracy, as they have a short period (epochs) to recover from damage in their structure.
%
Overall, the weight rewinding of LTH when the pruning removes filters and layers shares similar behavior.

The previous discussion takes into account the original setup by Frankle et al.~\shortcite{Frankle:2019} and its variations~\cite{Frankle:2020,Renda:2020}.
%
Liu et al.~\shortcite{Liu:2019} suggested an alternative to recover the weights for some initialization composing the training trajectory ($\theta_i$) of the dense network. The authors demonstrated that randomly re-initializing subnetworks allows them to become winning tickets. 
From the lens of layer pruning, we observe that subnetworks share the same trends. More specifically, on the rewinding by Liu et al.~\shortcite{Liu:2019}, our subnetworks outperform the dense network by up to $0.85$ percentage points.
%

Altogether, the previous results confirm the existence of winning tickets when the pruning process removes entire layers; thus answering our first research question. 



\noindent
\textbf{Picking Winning Tickets at Initialization.} So far, our experiments confirm the existence of winning tickets with the weight rewinding strategy, as in the traditional LTH conjecture. Additionally, we observe that relaxing the weight rewinding and re-initializing the subnetworks enable them to become winning tickets. Both settings assume that winning tickets emerge from a well-trained dense network.
In this experiment, we remove this constraint and propose to discover winning tickets without any training. In other words, we propose to find winning tickets before training begins.
Such a setting eliminates the requirement of heavy computational resources for training the initial (over-parameterized) dense network, an important step
towards democratization and green machine learning.
As we discussed before, early works have proposed to pick winning tickets at initialization~\cite{Lee:2019,Wang:2020,Tanaka:2020,Liu:2022}. However, these works focus on removing weights while we target removing layers, which is a more efficient and hardware-agnostic form of pruning.

One relevant question to find subnetworks from initialization is the choice of the criterion for estimating structure importance. It turns out that most criteria fail to measure importance without any training as the weights change drastically (hence the importance score). In summary, the criteria evaluated in Table~\ref{tab:rewind} are unsuitable\footnote{In fact, we can always employ the random criterion; however, we are interested in a systemic process for selecting winning tickets at initialization.} to prune at initialization.

To face the above issue, we employ two criteria specifically designed for estimating importance at initialization. The first criterion, named SNIP, computes the importance by multiplying the weight by its gradient (at the initial -- randomly -- configuration); then, it takes the absolute value of the resulting operation~\cite{Lee:2019}. The second criterion, termed GraSP, is a variant of SNIP, in which the importance considers the signal of the weight~\cite{Wang:2020}. 
Importantly, SNIP and GraSP are data-driven criteria since both forward data (and labels) through the network to compute the gradient during the importance estimation.
\begin{table}[!t]
	\centering
	\renewcommand{\arraystretch}{1.2}
	\caption{Predictive ability of tickets from ResNet32 when our LTH discovers them at initialization. 
	}
	\label{tab:wt_initialization}
	\begin{tabular}{ccc}
		\hline
		Pruning Density & SNIP & GraSP \\ \hline
		\multicolumn{1}{c|}{1} & (+) 0.08 & (+) 0.40 \\
		\multicolumn{1}{c|}{2} & (+) 0.00 & (+) 0.72 \\
		\multicolumn{1}{c|}{3} & (-) 0.50 & (+) 0.34 \\
		\multicolumn{1}{c|}{4} & (-) 0.53 & (+) 0.41 \\
		\multicolumn{1}{c|}{5} & (-) 1.00 & (+) 0.12 \\ \hline
	\end{tabular}
\end{table}

Table~\ref{tab:wt_initialization} summarizes the results. From this table, we highlight the following observations. (i) SNIP is an inefficient criterion for discovering very sparse subnetworks (tickets) since its tickets are no longer winning tickets after removing three layers. (ii) All tickets from GraSP win the lottery. Particularly, GraSP provides subnetworks with predictive ability superior to the dense network even on the highest pruning density we consider (5 layers -- 66\% of layers of ResNet32).
%

In order to confirm the superiority of Grasp over SNIP, we compare the subnetworks yielded by each criterion with the dense network through the lens of representation similarity among models.
For this purpose, we employ the method by Kornblith et al.~\shortcite{Kornblith:2019} (named CKA) that measures the representation similarity of two (architecturally equal or distinct) models. The idea is to verify if this similarity correlates with the results in Table~\ref{tab:wt_initialization}. On the lowest pruning severity ($p=1$), the difference in CKA similarity between the dense network and the subnetworks from GraSP and SNIP is less than one percentage point, with the subnetwork from GRASP obtaining the higher similarity. However, on the highest pruning severity ($p=5$), the subnetwork from SNIP exhibited a difference of almost two percentage points. 
On the one hand, the representation similarity between the subnetworks yielded by SNIP and the dense network decreases as a function of the pruning severity. On the other hand, according to the CKA similarity, the internal representation of subnetworks from GraSP remains (partially) undamaged compared to the dense network. This observation reinforces the positive results of GraSP in Table~\ref{tab:wt_initialization} and indicates that we can effectively extract winning tickets at initialization using GraSP.
%
Unless stated otherwise, we employ GraSP to discover winning tickets at initialization in the subsequent experiments.

\noindent
\textbf{Comparison with standard (structured) LTH.} This experiment compares our winning tickets (layer pruning) with the standard structured LTH (filter pruning). During this evaluation, we extract winning tickets of layer and filter pruning at initialization. 
Here, a critical aspect is how to produce subnetworks with the same sparsity (in terms of filters) when the pruning removes different structures. 
It turns out that when it eliminates layers, the sparsity becomes inflexible. For example, by removing 1 and 2 layers, the number of remaining filters is 1200 and 1168 and we cannot obtain a filter sparsity among these values (i.e., 1190).
For fairness of comparison, we adopt the following procedure. First, we remove layers from a dense network and calculate the number of filters in the obtained subnetwork (i.e., the number of kept filters). Then, to produce comparable subnetworks removing only filters, we run the pruning process forcing it to eliminate the closest number of filters of subnetworks from layer pruning.
Thereby, subnetworks yielded from a dense network without layers and filters are as close as possible in terms of filter sparsity. We highlight that the opposite process -- first removing filters and then layers -- impairs a comparable sparsity.

Table~\ref{tab:comparison_filter} shows the results in the lowest sparsity regime (one layer removed -- $p=1$). On both SNIP and GraSP criteria, our winning tickets at initialization yielded early winning tickets with better improvements. In particular, the subnetworks from the standard structured LTH become winning tickets only if we relax Equation~\ref{eq:LTH} by increasing the value $\xi$.
\begin{table}[!b]
	\centering
	\renewcommand{\arraystretch}{1.2}
	\caption{Comparison between tickets from the standard structured LTH (filter pruning) and our LTH (layer pruning). On both criteria for pruning at initialization, our subnetworks outperform their dense (ResNet32) counterpart. In other words, our tickets become winning tickets (they win the lottery). However, tickets from the standard structured LTH hardly become winning tickets.}
	\label{tab:comparison_filter}
	\begin{tabular}{ccc}
		\hline
		 & SNIP & GraSP \\ \hline
		Filter (Standard LTH) & (-) 0.79 & (-) 0.11 \\
		Layers (Ours LTH) & (+) 0.08 & (+) 0.40 \\ \hline
	\end{tabular}
\end{table}

Previous observations revealed the difficulty of extracting winning tickets at initialization and confirmed that it is no better than the standard \emph{training and prune} paradigm~\cite{Frankle:2021,Jorge:2021}; thus, corroborating the poor results of filter pruning in Table~\ref{tab:comparison_filter}. Our winning tickets (layer pruning) pose a different perspective for such an issue: filters cannot be the most effective structure in structured LTH.
Technically speaking, removing a layer consists of eliminating a group of filters (in particular, all filters) from a specific location, see Appendix~\ref{sec:app_technicaldetails}. Hence, it sounds like a counter-intuitive behavior that our winning tickets obtain accuracy superior to the standard LTH. Our results, however, are aligned with a body of studies that confirm the benefits of removing layers over other structures~\cite{Zhang:2022,Zhou:2022,Han:2022}.
To reinforce these claims, we measure the representation similarity between the models of Table~\ref{tab:comparison_filter} and the dense network, similarly as we performed in previous experiments. 
On this metric, our subnetworks exhibited a higher similarity than subnetworks from filter pruning, regardless of the importance criterion (SNIP or GraSP). Such values suggest that our winning tickets kept an internal representation more similar to the original dense network than the ones found by the standard structured LTH, which supports their highest accuracies in Table~\ref{tab:comparison_filter}.

Overall, we believe the findings above open a new direction for research in LTH: the influence of the structure taken into account during the pruning process.
%

\begin{table*}[!t]
	\centering
	\renewcommand{\arraystretch}{1.2}
	\caption{Performance gains when the pruning discovers winning tickets (layer pruning) at initialization. The values indicate the improvements (in percentage -- the higher, the better) of tickets regarding the dense network.}
	\label{tab:computational_cost}
	\begin{tabular}{cccccc}
		\hline
		&  \begin{tabular}[c]{@{}c@{}}Pruning Density\\ (p)\end{tabular} & FLOPs & \begin{tabular}[c]{@{}c@{}}Memory\\ Consumption\end{tabular} & \begin{tabular}[c]{@{}c@{}}CO$_2$ \\ Emission\end{tabular} & \begin{tabular}[c]{@{}c@{}}Training\\ Speed up\end{tabular} \\ \hline
		\multirow{3}{*}{\begin{tabular}[c]{@{}c@{}}Winning Tickets\\ at Initialization\\ (ResNet32)\end{tabular}} & 1 & 6.82 & 10.13 & 11.11 & 1.10 \\
		& 3 & 20.47 & 25.29 & 25.92 & 1.35 \\
		& 4 & 34.13 & 30.11 & 33.33 & 1.57 \\ \hline
		\multirow{4}{*}{\begin{tabular}[c]{@{}c@{}}Winning Tickets\\ at Initialization\\ (ResNet56)\end{tabular}} & 3 & 11.25 & 15.15 & 13.33 & 1.16 \\
		& 4 & 15.00 & 19.42 & 20.00 & 1.24 \\
		& 5 & 18.76 & 23.69 & 24.44 & 1.34 \\
		& 12 & 45.00 & 45.82 & 51.11 & 2.03 \\ \hline
	\end{tabular}
\end{table*}
\noindent
\textbf{Computational Cost of Winning Tickets at Initialization.}
The aforementioned discussion confirms the existence of winning tickets at initialization when pruning removes layers (recall that our subnetworks come from layer pruning -- see Algorithms~\ref{alg::pruning} and~\ref{alg::early_ticket}). 
From a practical perspective, identifying winning tickets at initialization means that we can speed up the training stage and save computation in terms of different performance metrics, as we can replace the learning of a dense network with its sparse version. 
Such achievements become possible since we discover subnetworks that (i) are more efficient than the heavy and over-parametrized dense network; (ii) match the predictive ability of their dense equivalent (i.e., the subnetworks are winning tickets) and (iii) spend no additional cost since we discover them before any training epoch.

Table~\ref{tab:computational_cost} reports the performance gains of our winning tickets at initialization on standard pruning metrics such as floating point operations (FLOPs), memory consumption and training speed up. Following a modern trend~\cite{Lacoste:2019,Strubell:2019,You:2020}, we also report the CO$_2$ emission during the training stage. According to Table~\ref{tab:computational_cost}, it is evident the advantages of discovering winning tickets at initialization. For example, our winning tickets speed up the training time of ResNet32 and ResNet56 by up to $1.57\times$ and $2.03\times$, meaning that we can significantly reduce the costs involved during the training step such as energy consumption and CO$_2$ emission. Importantly, all these improvements come at no additional cost.

Previous attempts have proposed to find winning tickets early in the training phase~\cite{You:2020,Chen:2022}. Our work differs from these works, as we discover winning tickets prior to any training (roughly speaking, our subnetworks are the
earliest as possible), which is a more efficient strategy since the gains in Table~\ref{tab:computational_cost} occur before the training starts.

\noindent
\textbf{Robustness of Winning Tickets at Initialization.} There is a growing body of work evaluating pruning methods in adversarial images since the technique emerges as a powerful and efficient defense mechanism against adversarial attacks and out-of-distribution examples~\cite{Diffenderfer:2021,Liu:2022,T:2022,Chen:2022}. 
In this experiment, we demonstrate if there exists some sensibility of our winning tickets to these scenarios. For this purpose, we employ the CIFAR-C~\cite{Hendrycks:2019} and CIFAR-10.2~\cite{Lu:2020} datasets. It is important to mention that our training stage employs the standard clean training on CIFAR-10, which means that we do not employ any defense mechanism (i.e., adversarial training).

Table~\ref{tab:robustness} (third column) shows the improvements in robustness of winning tickets at initialization (the ones in Table~\ref{tab:computational_cost}) over the deep network used to discover them. According to the results, our winning tickets effectively increase robustness to adversarial images. Specifically, our winning tickets outperformed their respective dense network by up to $1.61$ and $3.73$ pp for ResNet32 and ResNet56, respectively. Intriguingly, for each architecture, one winning ticket exhibited low robustness, in which the highest decline is only $0.16$ pp.
\begin{table}[!b]
	\centering
	\renewcommand{\arraystretch}{1.2}
	\caption{Robustness against adversarial and out-of-distribution examples of our winning tickets at initialization. 
		On both CIFAR-C and CIFAR-10.2 datasets, we report the improvement (in percentage points -- the higher, the better) of subnetworks regarding their dense counterpart.}
	\label{tab:robustness}
	\begin{tabular}{cccc}
		\hline
		& p  & CIFAR-C & CIFAR-10.2 \\ \hline
		\multirow{3}{*}{\begin{tabular}[c]{@{}c@{}}Winning Tickets\\ at Initialization\\ (ResNet32)\end{tabular}}  & 1  & (+) 1.13                                                & (+) 0.00                                                        \\
		& 3  & (-) 0.05                                                & (-) 1.15                                                        \\
		& 4  & (+) 1.61                                                & (-) 0.65                                                        \\ \hline
		\multirow{4}{*}{\begin{tabular}[c]{@{}c@{}}Winning Tickets\\ at Initialization \\ (ResNet56)\end{tabular}} & 3  & (+) 0.59                                                & (+) 0.70                                                        \\
		& 4  & (+) 3.73                                                & (+) 1.75                                                        \\
		& 5  & (-) 0.16                                                & (+) 0.70                                                        \\
		& 12 & (+) 1.98                                                & (-) 1.95                                                        \\ \hline
	\end{tabular}
\end{table}

Regarding the out-of-distribution generalization (last column in Table~\ref{tab:robustness}), our winning ticket increase the generalization of ResNet56 in up to $1.75$. On the ResNet32, however, we observe that the results become negative as a function of the pruning severity; thus raising the question of whether our tickets generalize well in out-of-distribution examples when extracted from shallow architectures. The analysis of such behavior is an interesting direction for future work.

%
For some settings (5 out of 14 configurations), our winning tickets underperform their dense equivalent. We did not observe any pattern in these low-performance tickets and left an in-depth investigation of this issue for future research.  
Overall, the achievements in robustness against adversarial images and out-of-distribution examples suggest that the benefits of our winning tickets are beyond reducing computational performance, which aligns with findings of previous works: removing structures from networks increases robustness and out-of-distribution generation~\cite{Diffenderfer:2021,Chen:2022,T:2022}.

%% file: Sections/Conclusions.tex
\section{Conclusions}\label{sec:conclusions}
In this work, we explore the concepts of the Lottery Ticket Hypothesis (LTH) and pruning at initialization from the lens of layer pruning. From this form of pruning, we bring both theoretical and practical benefits.
First, we verify the behavior of LTH when the pruning yields subnetworks (tickets) by removing layers from a dense network, in which we confirm that the tickets successfully become winning tickets.
%
Built upon this, we propose to systematically discover winning tickets at initialization, which means identifying sparse subnetworks (from layer pruning) without the need of training a dense network.
According to extensive experiments, our winning tickets at initialization speed up the learning phase by up to $2\times$, reducing the carbon emission by up to $51\%$.
Such achievements come at no additional price, as we extract winning tickets before training the dense network. Additionally, our winning tickets become more robust against adversarial images and generalize better in out-of-distribution examples than their dense counterpart. 
%
%
We hope these benefits attract more research on LTH from the lens of layer pruning.

\noindent
\textbf{Limitations.}
Our strategy that discovers winning tickets at initialization employs well-established criteria for assigning the importance of each layer before training (i.e., using the randomly
initialized parameters to guide the pruning algorithm). Hence, it inherits the limitation of being unable to run on an iterative pruning scheme, which has been demonstrated to be more effective than one-shot pruning. Due to theoretical and technical details involving layer pruning, our structured LTH is confined to the realm of residual-like architectures. Thus, our LTH is infeasible to plain networks (e.g., VGG-16), as well as other works of this category of pruning. Fortunately, most modern architectures employ some type of residual connection.

\noindent
\textbf{Open Questions.} 
Throughout the experiments, we observe that our subnetworks became winning tickets more easily than the ones yielded by filter pruning. Such a finding raises the question of whether the removed structure matters in the success of LTH at initialization.
Aligned with this issue, an exciting direction for future investigation is studying the behavior of LTH when the pruning algorithm removes multiple structures (weights, filters and layers) simultaneously as well as how to prune these structures together at initialization.
%

%% file: Sections/Acknowledgments.tex
\section*{Acknowledgments}
The authors would like to thank CNPq (grant \#309330/2018-1).

%% file: Sections/Appendix.tex
\appendix\section{Appendix}\label{sec:appendix}
\subsection{Layer Pruning}\label{sec:app_technicaldetails}
\noindent
\textbf{Theoretical Issues.}
Assume a network $\mathcal{F}$ of $L$ layers as a set of $L$ transformations $f_i(.)$. For the sake of simplicity, $f_i$ consists of a series of convolution, batch normalization, and activation operations. In this definition, we obtain the network output ($y$) by forwarding the input data through the sequential layers $f$, where the input of layer $i$ is the output of the previous layer $i-1$; therefore, $y = f(x)$ $f_L(...f_2(f_1(.)))$.
This composes the idea behind plain networks (i.e., VGG). 

In residual-like networks, the output of layer $i$, $y_i$, consists of the transformation $f_i$ plus the input it receives $y_{i-1}$ (see Figure~\ref{fig:residualmodule}). Formally, we can define the output of the $i$-th layer as
\begin{equation}\label{eq:residual}
	y_i = f_i(y-1) + y_{i-1}.
\end{equation}
Equation~\ref{eq:residual} composes a residual module, where the rightmost part is named \emph{identity-mapping shortcut} (or identity for short). It is important to observe in Equation~\ref{eq:residual} that if we disable $f(i)$ (a layer) then $y_i = y_{i-1}$.

Veit et al.~\shortcite{Veit:2016} showed that the identity enables the information to take different paths in the network, in the sense that, we can disable some $f_i$ without degrading (or with negligible damage) the expected representation of the subsequent layers (i.e., $f_{i+1}$). In other words, some layers $f_i$ do not depend strongly on each other; hence, we can eliminate them. For example, in Figure~\ref{fig:residualmodule}, we could remove layer $i$ with no loss in the predictive ability of the network. On the other hand, due to the absence of identity, plain networks meet collapse in the representation if we remove only one of their layers. 
We refer interested readers to Figure 3 of the study by Veit et al.~\shortcite{Veit:2016} for a comparison of accuracy drop between residual and plain networks.

\begin{figure}[!htb]
	\centering
	\includegraphics[width=\columnwidth]{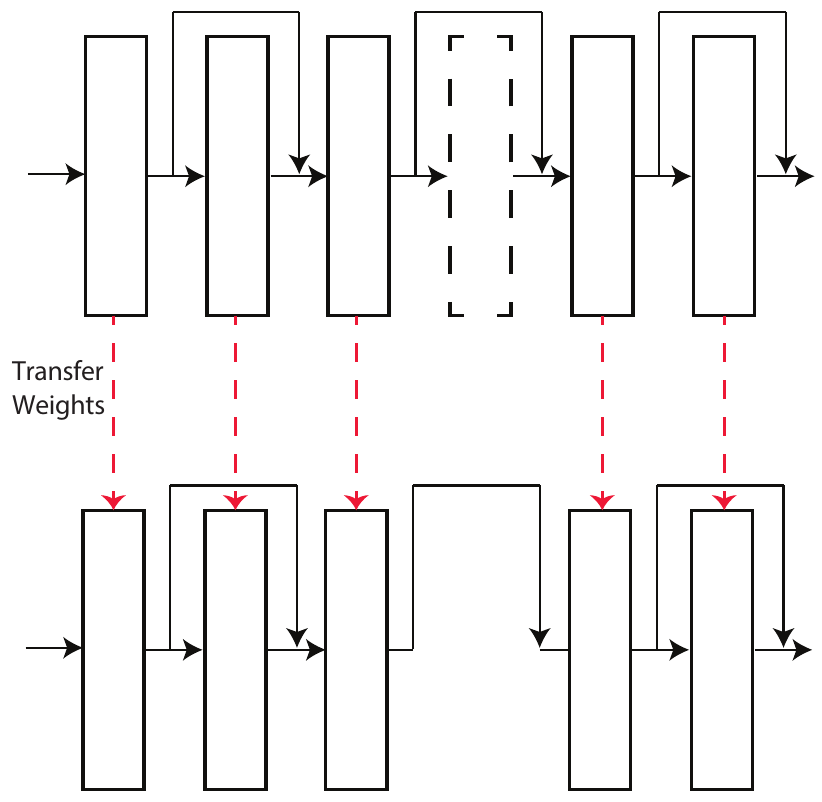}
	\caption{Overall process to remove layers (residual models) from a residual network. After identifying a victim layer (dashed rectangle), we create a novel network (bottom) without it. Finally, we transfer the weights (red arrows) of the kept layers from the original unpruned network (top) to the new network.}
	\label{fig:weighttransfer}
\end{figure}

\noindent
\textbf{Technical Issues.} We can disable layer $i$ by setting its weighs to zero (the widely employed zeroed-out scheme). This way, the output of layer $i-1$ is directly connected to layer $i+1$ (see Figure~\ref{fig:residualmodule} middle). 
However, such a process does not achieve performance gains without specialized frameworks or hardware for sparse computation. Instead of zeroing weights, we can perform the following process.
After identifying which layers remove (i.e., a victim), we create a new network without layer $i$ and transfer the weights of the kept layers to the new network. For example, if we have a network with $L$ layers and want to remove $k$ layers, then, we create a novel network with $L-k$ layers. In summary, the pruned network (bottom in Figure~\ref{fig:weighttransfer}) inherits the weights of the kept layers of the original network (top in Figure~\ref{fig:weighttransfer}).

We highlight that the pruning cannot remove some layers due to incompatible dimensions of (input/output) tensors. Such an incompatibility comes from the spatial resolution layer (downsampling layers). More specifically, we cannot remove layers before and after the downsampling layers. Importantly, filter pruning also suffers from this issue.
%
%
%

\subsection{$\ell_1$-norm}\label{sec:l1_histogram}
\begin{figure}[!htb]
	\centering
	\includegraphics[width=\columnwidth]{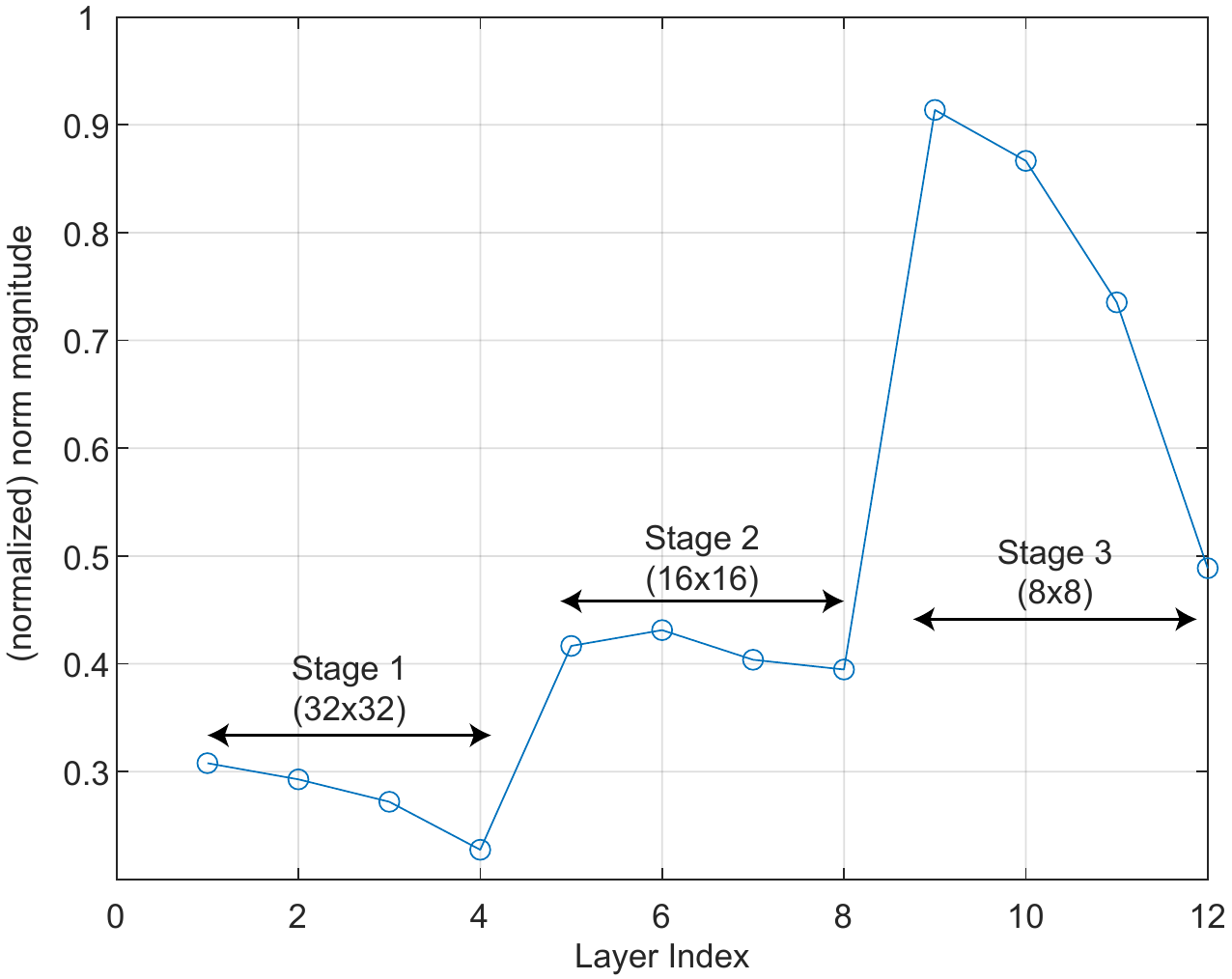}
	\caption{$\ell_1$-norm score of layers of ResNet32. Layers within a stage operate on the same input/output spatial resolution (i.e., the size of the feature map -- values in parentheses).}
	\label{fig:l1}
\end{figure}

\begin{figure*}[ht!]
	\centering
	\includegraphics[width=\textwidth]{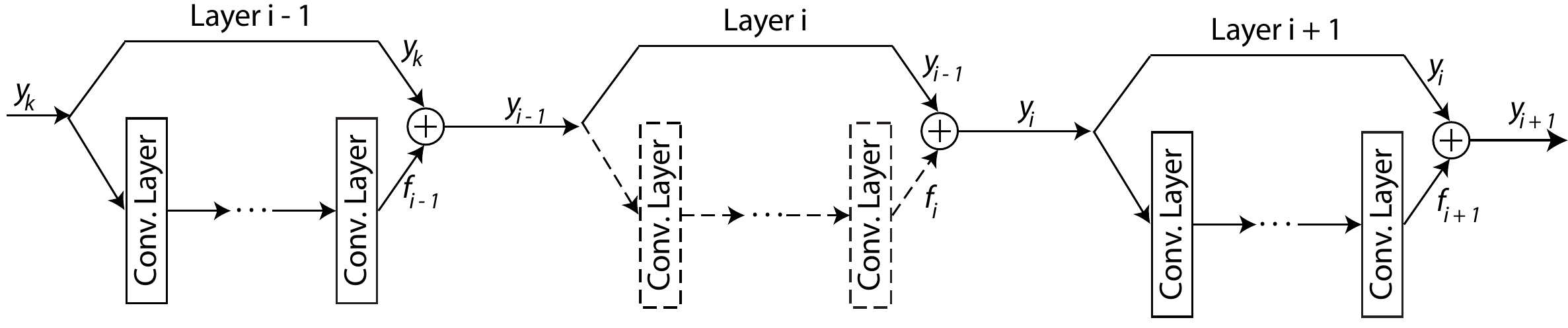}
	\caption{Architecture of a residual-like network. The rationale behind this architecture is that the output of a layer takes into account the transformation performed by it ($f$) plus ($\oplus$) the input ($y$) it receives. Due to this essence, when we disable layer $i$ (its transformation -- dashed lines), the output (representation) of layer $i-1$ is propagated to layer $i+1$, which means that the output $y_i$ belongs $y_{i-1}$. For the sake of simplicity, we omit the batch normalization and activation layers.}
	\label{fig:residualmodule}
\end{figure*}

Figure~\ref{fig:l1} illustrates the $\ell_1$-norm scores of layers (the ones that the pruning could remove) of ResNet32. From this figure, we see that the magnitude of scores of layers correlates with the stage (groups of layers operating on the same resolution of feature maps) to which they belong.
Since our pruning strategy takes into account all layers (i.e., all scores) at once, this criterion is infeasible. Specifically, there is a bias to layers of early stages (i.e., they will always be selected as victims).
